\documentclass[acmtog, nonacm]{acmart}

\AtBeginDocument{%
  \fancypagestyle{plain}{%
    \fancyhf{} %
    \fancyfoot[L]{ACM SIGENERGY Energy Informatics Review}%
    \fancyfoot[R]{Volume 5 Issue 2, July 2025}}
  \providecommand\BibTeX{{%
    \normalfont B\kern-0.5em{\scshape i\kern-0.25em b}\kern-0.8em\TeX}}}
    
\let\oldmaketitle\maketitle
\renewcommand{\maketitle}{%
  \oldmaketitle%
  \thispagestyle{plain}%
  \pagestyle{plain}}

\setcopyright{acmcopyright}
\copyrightyear{2025}
\acmYear{2025}
\acmDOI{XXXXXXX.XXXXXXX}
\acmConference[HotCarbon '25]{Noise-aware Client Selection for carbon-efficient Federated
  Learning via Gradient Norm Thresholding}{July 10--11,
  2025}{Boston, MA}
\acmPrice{15.00}
\acmISBN{978-1-4503-XXXX-X/18/06}

\usepackage{rotating}

\usepackage{etoolbox}
\usepackage{float}
\usepackage{tabularx}  
\usepackage{comment}
\usepackage{xcolor}
\usepackage{todonotes}
\usepackage[utf8]{inputenc}
\usepackage[linesnumbered,ruled,vlined]{algorithm2e}
\usepackage{xcolor}

\definecolor{teal}{RGB}{0,128,128}

\SetCommentSty{mycommfont}
\SetAlFnt{\ttfamily}
\SetKwComment{Comment}{$\triangleright$\ }{}

\definecolor{ineseGreen}{rgb}{0.6, 1.0, 0.2}

\usepackage{soul,xcolor}
\definecolor{mintgreen}{rgb}{0.74, 0.98, 0.78}  
\sethlcolor{mintgreen}

\SetKwInOut{Input}{Input}
\SetKwInOut{Output}{Output}
\usepackage{enumitem}
\usepackage{subcaption}
\usepackage{float}

\usepackage[table]{xcolor}
\definecolor{lightgray}{gray}{0.9}
\definecolor{lightblue}{rgb}{0.85,0.92,1} 
\begin{document}

\title{Noise-aware Client Selection for carbon-efficient Federated Learning via Gradient Norm Thresholding}

\author{Patrick Wilhelm}
\email{patrick.wilhelm@tu-berlin.de}
\affiliation{%
  \institution{BIFOLD}
  \institution{Technische Universität Berlin}
  \city{Berlin}
  \country{Germany}
}

\author{Inese Yilmaz}
\email{inese.yilmaz@campus.tu-berlin.de}
\affiliation{%
  \institution{Technische Universität Berlin}
  \city{Berlin}
  \country{Germany}
}

\author{Odej Kao}
\email{odej.kao@tu-berlin.de}
\affiliation{%
  \institution{Technische Universität Berlin}
  \city{Berlin}
  \country{Germany}
}


\begin{abstract}
Training large-scale Neural Networks requires substantial computational power and energy. Federated Learning enables distributed model training across geospatially distri\-buted data centers, leveraging renewable energy sources to reduce the carbon footprint of AI training. Various client selection strategies have been developed to align the volatility of renewable energy with stable and fair model training in a federated system. However, due to the privacy-preserving nature of Federated Learning, the quality of data on client devices remains unknown, posing challenges for effective model training.
In this paper, we introduce a modular approach on top to state-of-the-art client selection strategies for carbon-efficient Federated Learning. Our method enhances robustness by incorporating a noisy client data filtering, improving both model performance and sustainability in scenarios with unknown data quality. Additionally, we explore the impact of carbon budgets on model convergence, balancing efficiency and sustainability. Through extensive evaluations, we demonstrate that modern client selection strategies based on local client loss tend to select clients with noisy data, ultimately degrading model performance. To address this, we propose a gradient norm thresholding mechanism using probing rounds for more effective client selection and noise detection, contributing to the practical deployment of carbon-efficient Federated Learning.

\end{abstract}


\received{19 Mai 2025}

\maketitle

\section{Introduction}

The rapid advancement of deep learning has led to models with hundreds of billions of parameters, consistently outperforming previous AI benchmarks. However, this progress comes at a significant cost: larger models demand more computational power, resulting in hours of GPU training time, increased energy consumption, and carbon emissions of up to 24.7 tonnes of CO\(_2\) for training a model with approximately 170 billion parameters \cite{luccioni2023estimating}. Recent studies project that the energy demand of global data centers will reach 1,000 TWh by 2026, driven in large part by the surge in AI inference and training workloads. Furthermore, carbon emissions from these activities are estimated to contribute up to 8\% of global emissions within the next decade \cite{designingSustainableCompSystems}.

One of the most promising strategies to mitigate the carbon footprint of neural network training is to align computational workloads with the availability of renewable energy sources~\cite{patterson2022carbon, sukprasert2024limitations}. Distributed training approaches, particularly those leveraging geospatially distributed data centers, offer a unique opportunity to utilize local renewable energy surpluses. Federated Learning (FL), in particular, enables decentralized training across data centers or edge devices, making it a suitable candidate for carbon-aware, geospatial model training \cite{wiesner2024fedzero, bian2024cafe}.

To exploit renewable availability, client selection strategies in FL have been developed to dynamically allocate training workloads based on the local computing capacity and renewable energy availability of clients \cite{carbonaware_scheduling_hotcarbon24}. However, due to the privacy-preserving nature of FL, the quality of local data remains unknown. This presents a challenge: current selection strategies often rely on client-side training loss to determine utility, yet high loss may stem from either valuable, hard examples—or noisy and corrupted data \cite{oort}. Consequently, selecting clients solely based on high loss can inadvertently introduce harmful noise into the collaborative training process.

Achieving a balance between sustainability and model performance requires more nuanced approaches that can infer the impact of a client's data without breaching privacy. We propose a noise-aware client selection mechanism that leverages gradient norm statistics as a proxy for data quality. Inspired by the Critical Learning Periods \cite{achille2017critical, criticalFL}, we show that estimating the Fisher Information Matrix via gradient norm during probing rounds allows for identifying clients whose data positively contribute to model convergence—while filtering those with potentially corrupted or noisy samples.

In our experiments, we demonstrate the following key findings:
\begin{itemize}
    \item Enhancing traditional loss-based client selection with a single gradient norm thresholding round improves resilience against noisy or corrupted client data.
    \item Integrating carbon budgets into the client selection process enables a more balanced trade-off between model performance and sustainability in carbon-aware FL.
\end{itemize}

Our findings are supported by a series of experiments, presented throughout the remainder of this paper. In Section~\ref{sec:relatedwork}, we discuss related work on carbon-aware FL, noise-robust training and data valuation. Section~\ref{sec:system_design} outlines our system design, including carbon data integration and our proposed client selection approach. In Section~\ref{sec:experiments}, we evaluate the effectiveness of gradient norm thresholding for filtering noisy clients.  Finally, in Section~\ref{sec:discussion} and ~\ref{sec:conclusion}, we summarize our contributions and provide an outlook on incorporating data valuation techniques into privacy-preserving FL.

\begin{figure*}[h!]
    \centering
    \includegraphics[trim=1.1cm 14cm 7.7cm 2cm, clip, width=0.7\textwidth]{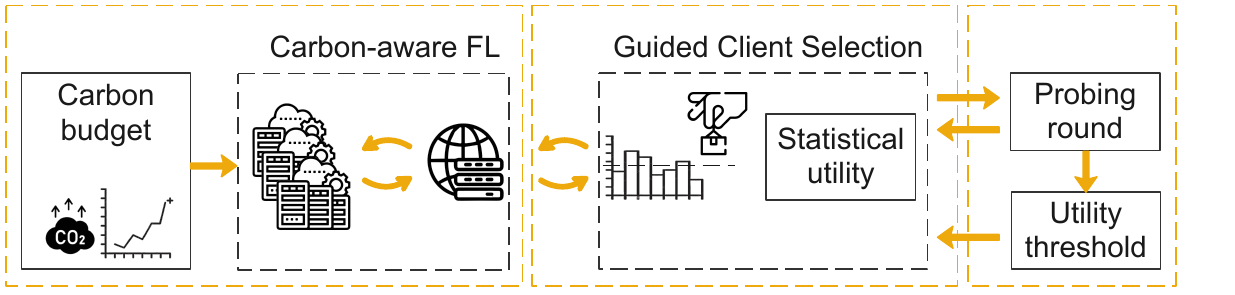}
\caption{System Overview: Clients represent datacenters in 30 regions within the US. While the data quality is unknown we introduce probing rounds in the beginning of federated training to identify clients with noisy data. Integrating carbon budgets we can balance clients with high data utility and their current carbon intensity.}
    \label{fig:objectives}
\end{figure*}

\section{Related Work and Background}
\label{sec:relatedwork}

Previous studies align client selection strategies with the high volatility of renewable energy within the power grid \cite{bian2024cafe, wiesner2024fedzero}. Recent studies have investigated various strategies for client selection, including choosing clients according to the size of their local datasets \cite{mcmahan2017communication}, or favoring clients that exhibit higher training losses \cite{cho2020client, oort}. None of those take data quality into account.

Noisy data can significantly disrupt the training process by producing high
loss and erratic gradients, leading to unstable parameter updates and reduced learning efficiency \cite{FL_poisoning_filter_by_validationset}. While examples with slightly higher losses are valuable for guiding meaningful learning, excessively large training losses often indicate noisy or corrupted data that should be avoided \cite{GUPTA2019466, li2020learningdetectmaliciousclients, FL_noisy_clients, FL_poisoning_filter_by_validationset}.  

Critical learning periods in neural networks are early phases of training during which the model is most plastic and responsive, forming lasting internal representations that significantly influence its final performance \cite{golatkar2019time, achille2017critical}.
During this phase, distinguishing between informative, high-impact examples and harmful noisy data, enables
more effective and stable training \cite{kleinman2023critical}.
Several works approximate the Hessian as an indicator for critical learning periods using Fisher Information Matrix \cite{yan2021critical, jastrzkebski2018relation}, which by itself can be approximated via gradient norm for more efficient computation \cite{criticalFL}. 

\section{System Design and Approach}
\label{sec:system_design}

In Federated Learning (FL), data is typically non-IID across clients, leading to significant variations in data quality and utility. Some clients may hold high-quality, diverse, and representative datasets that support convergence and generalization, while others may hold noisy, biased, or redundant data that hinders training and degrades overall performance.

In our system, we extend client selection strategies through an explicit probing round at the beginning and consecutive filtering via a coefficient threshold to maintain model quality and avoid selecting harmful data.

While aligning computation with renewable energy availability promotes sustainability, it restricts the client pool and introduces variability in participation. This often leads to reliance on suboptimal clients, negatively affecting training efficiency and model accuracy. While the use of a limited carbon budget increases client availability, it is essential to prioritize clients not only based on energy availability but also on their expected contribution to model performance.

We enable the selection of higher-emission clients through the use of a fixed carbon budget and improve budget efficiency via informed client selection. To achieve this, we model the trade-off between clients’ statistical utility and carbon intensity by adapting Oort’s reward calculation mechanism  \cite{oort}.
Figure \ref{fig:objectives} visualizes our noise-aware client selection extension for carbon-aware FL.

\subsection{Client Selection through Gradient Norm Probing}
\label{sec:oortwt}

To enable robust client selection under real-world data noise, we begin by evaluating all clients during an initial probing round. Clients are assessed using the statistical utility formulation proposed in Oort~\cite{oort}, but instead of using the loss approximation, we compute utility based on the gradient norm once at the beginning: \( U(i) = |B_i| \cdot \sqrt{\frac{1}{|B_i|} \sum_{k \in B_i} \|\nabla f(k)\|^2} \), where \( \nabla f(k) \) is the gradient of the loss function with respect to sample \( k \), and \( \|\nabla f(k)\| \) denotes its L2 norm. \(U_i\) is the utility and \(B_i\) the local data of client \(i\).

This utility, further referred to as probing utility, reflects the curvature of the local loss landscape, capturing data informativeness and noise sensitivity more effectively than training loss. While more computationally intensive, it offers greater robustness in federated settings. 

The server aggregates client probing utilities and applies a utility-variance-based threshold. Specifically, a client is retained if its probing utility satisfies \( \text{utility} \geq c \cdot \max(\text{utility}) \), where \( c \in [0,1] \) is a configurable coefficient that controls the exclusion level, and its effect is evaluated in ~\ref{sec:exp1}. This ensures that only the highest-utility clients are retained, while those falling below the threshold are excluded from subsequent training rounds.

This approach is not limited to the gradient norm–based utility examined in our study; other metrics can be used in its place or combined to evaluate clients during the probing round. Different thresholding strategies can then be applied to enable more robust client selection. Moreover, the overall method is compatible with other client selection approaches.

\subsection{Utility Aware Carbon Budget Allocation}
\label{sec:oortca}

To address the limitations of energy-aligned training—such as reduced client availability and reliance on suboptimal participants—we explore a client selection method that accounts for a fixed carbon budget and allocates resources based on client utility. Our method builds on Oort’s client selection strategy~\cite{oort}, enhancing it with carbon-awareness to improve training efficiency without exceeding emissions constraints.

We replace Oort’s exploration mechanism with the probing round introduced earlier, during which we can also compute statistical utility as proposed in Oort\cite{oort} for all clients. Rather than selecting only the top-scoring clients based on utility alone, we formulate a budget-aware optimization problem. In each round, the server selects a subset of clients that maximizes the total utility score while ensuring that the round’s total emissions remain within the carbon budget:
\noindent
\[
\max_{S \subseteq \mathcal{C}} \sum_{i \in S} r_i \quad \text{s.t.} \quad \sum_{i \in S} c_i \leq B_t, \quad |S| \leq K
\]
\noindent
Here, \( \mathcal{C} \) is the set of all clients, \( r_i \) is the utility score of client~\(i\) (obtained from the probing round), \( c_i \) is the carbon intensity of client~\(i\) in round~\(t\), \( B_t \) is the carbon budget allocated for round~\(t\), and \( K \) is the number of clients selected per round.

By explicitly modeling the trade-off between utility and emissions, this strategy improves budget efficiency and mitigates the performance degradation often caused by over-reliance on low-emission clients.

\begin{figure}[h!]
  \centering
  \includegraphics[width=\linewidth]{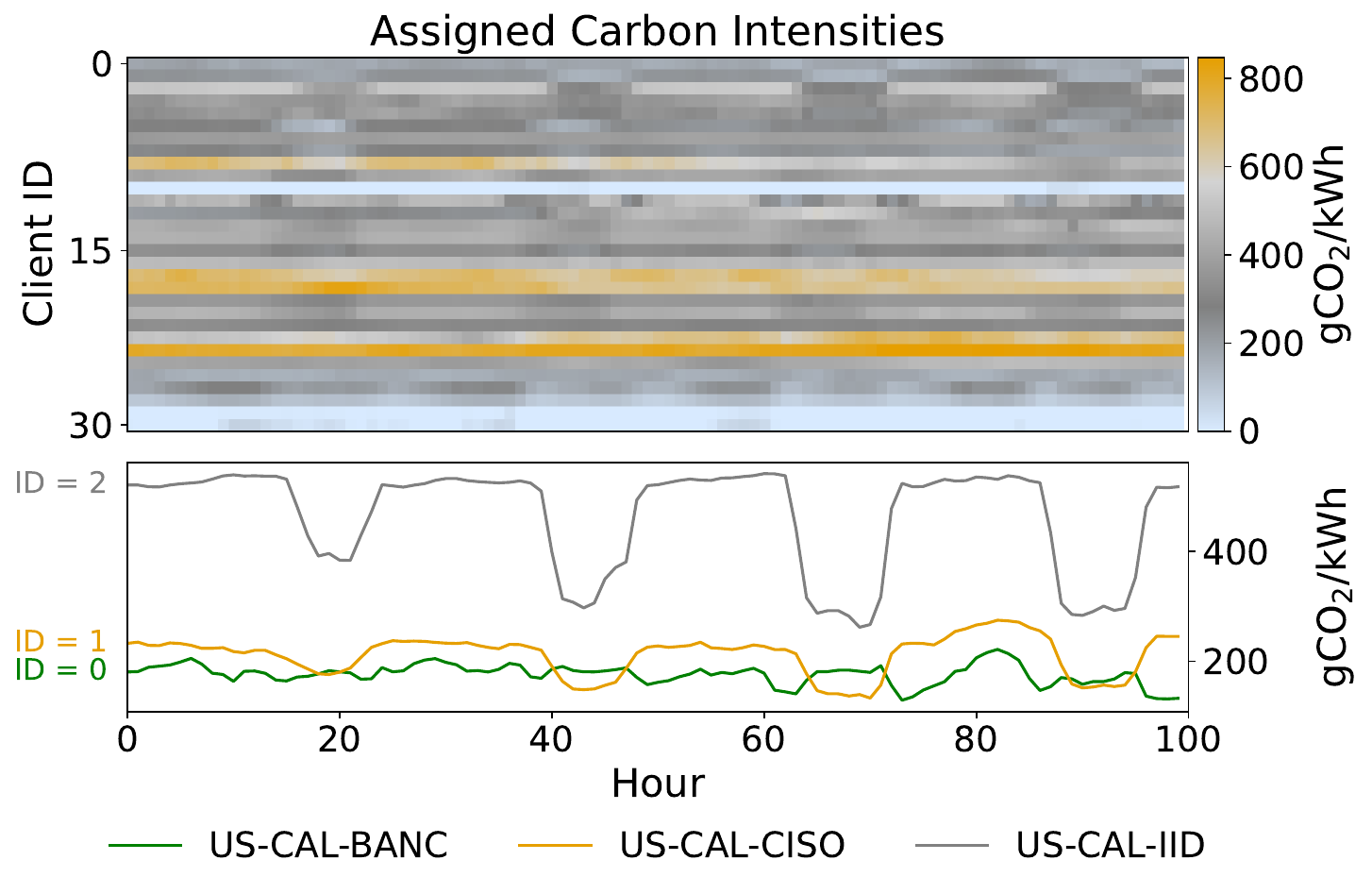}
  \caption{Assigned carbon intensity values for clients over 100 training rounds.
Each round corresponds to one hour, spanning from \texttt{2023-01-15 00:00:00} to \texttt{2023-01-19 04:00:00} (UTC). The bottom plot highlights daytime intensity patterns for three clients from different regions.}
  \label{fig:carbonintensity}
\end{figure}

\section{Experiments}
\label{sec:experiments}

We evaluate our proposed strategies under two practical considerations: data quality variation and the temporal–spatial variability of carbon intensity across energy regions. 

We perform experiments using CIFAR-10 \cite{cifar10_krizhevsky2009learning} with data distributed across clients non-IID under Dirichlet ($\alpha=10$) on 30 clients. A simple CNN (2 convolutional and 3 fully connected layers) is trained with 10 clients per round, 2 local epochs, batch size 32, using Adam optimizer (learning rate = 0.001). 

To assess robustness to data noise, specifically inaccuracies or inconsistencies in input features that disrupt the relationship between features and their corresponding labels, we simulate feature noise by adding zero-mean Gaussian noise with standard deviation \( \sigma = 1 \) to each image, followed by restricting pixel values to the valid range \([0, 1]\). This perturbs image details while slightly preserving the overall structure. We replace the data of 6 clients with corrupted, noisy variants. 

To model and assess how carbon intensity data can be utilized in combination with statistical utility, each of the 30 clients is assigned to a U.S. energy region, using hourly carbon intensity historical data from Electricity Maps~\cite{electricitymaps2024}, as illustrated in Fig.~\ref{fig:carbonintensity}. For simplicity, we assume that all clients have equal power demand, with each client consuming 1kWh during training, and each training round lasting one hour.

For carbon budgeting, energy curtailment data~\cite{electricitymaps2024} is combined with carbon intensity data: if curtailment is available, the client is assumed to operate on curtailed (zero-emission) energy and assigned an intensity of 0; otherwise, the previously recorded intensity value is used.

\subsection{Handling Noisy Clients via Probing-Based Client Evaluation}
\label{sec:exp1}

We evaluate gradient norm thresholding, described in Section~\ref{sec:oortwt}, on a data distribution that includes six corrupted clients, by applying it to two standard baselines: Random and Oort~\cite{oort}. In this configuration, we use only Oort’s statistical utility for its reward mechanism, excluding system utility to focus specifically on data quality. The resulting thresholded variants are referred to as RandomWT and OortWT, respectively.

\begin{figure}[h!]
    \centering
    \includegraphics[width=0.48\textwidth]{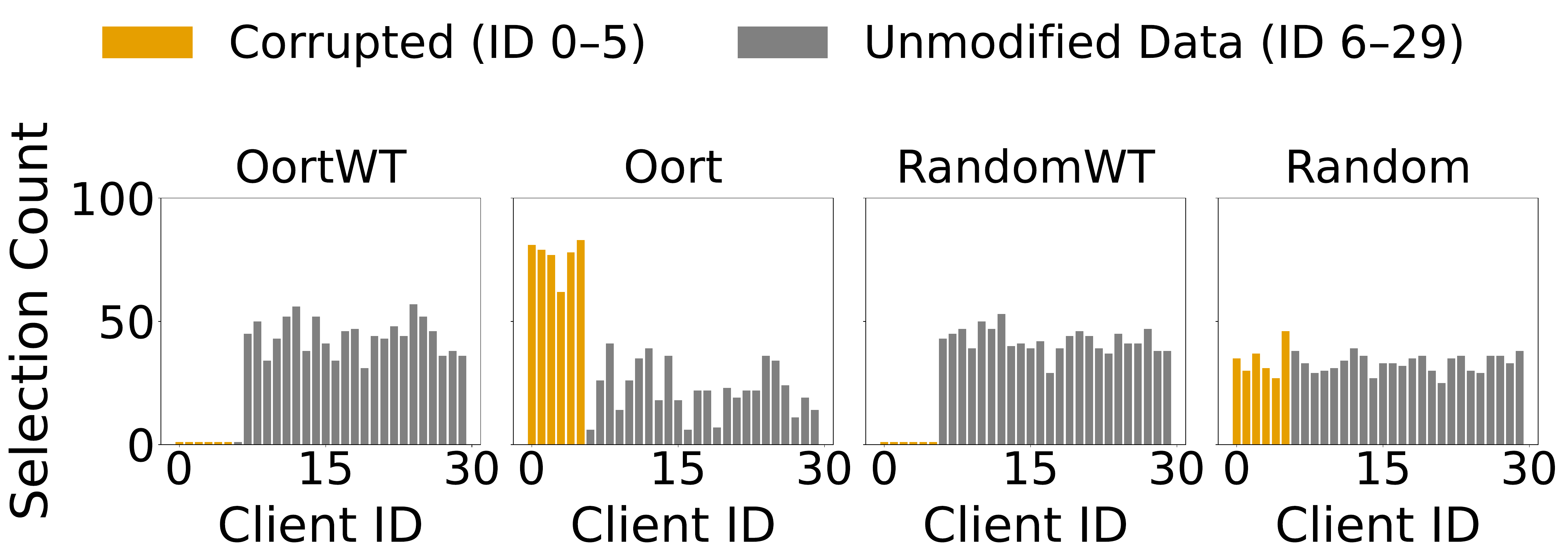}
    \caption{Client selection counts, 6 out of 30 clients contain noisy data.}
    \label{fig:noisyclients}
\end{figure}

Without thresholding, corrupted clients are frequently selected for training. As shown in Figure~\ref{fig:noisyclients}, Oort consistently prioritizes these clients due to its utility metric, which is based on local training loss—typically increased for noisy data. Although Oort includes an optional blacklisting mechanism that allows the exclusion of clients after participating in a specified number of rounds, it continues to favor corrupted clients prior to that point. 
In contrast, gradient norm thresholding enables the identification of noisy clients and their exclusion of subsequent training rounds.

\begin{figure}[b]
    \centering
    \includegraphics[width=0.4\textwidth]{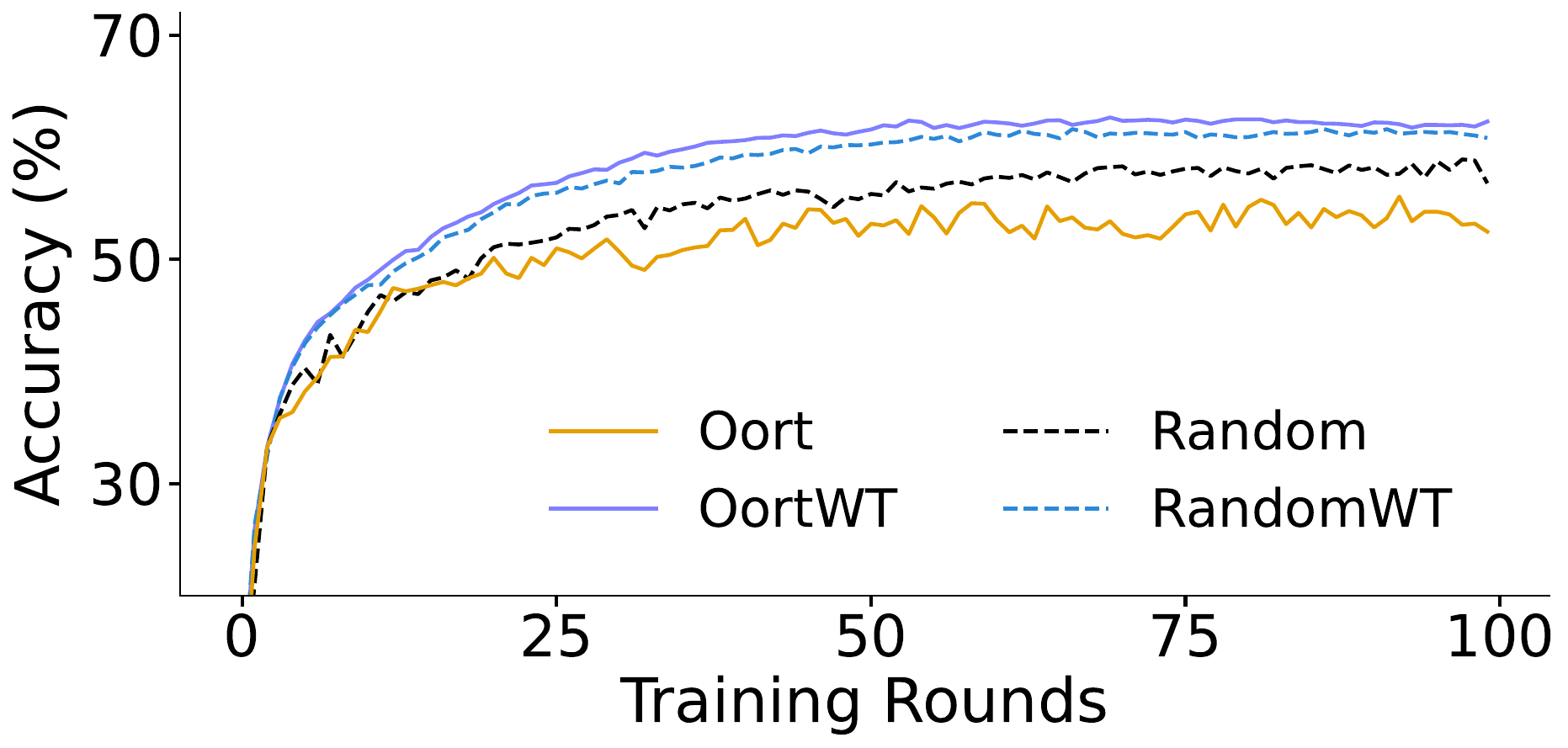}
  \caption{Convergence curves under noisy data scenario. 
  Baseline methods are compared with their thresholded variants. Thresholding reduces the impact of corrupted clients, resulting in faster and more stable convergence.}
  \label{fig:convergence}
\end{figure}

Selecting an appropriate threshold is key to filtering out noisy or harmful clients while retaining useful ones. In our setup, a value of \( c = 0.5 \) provides a good balance, in practice, this would require evaluation on a small, representative subset of the data.


The results in Figure~\ref{fig:convergence} demonstrate that the overselection of noisy clients significantly degrades Oort’s performance, leading to reduced accuracy. While Random selection does not systematically favor noisy clients, it still includes them with high probability, which disrupts convergence. In contrast, the thresholded variants, OortWT and RandomWT, result in faster and more stable convergence and achieve higher final accuracy. The improvement is particularly beneficial for methods like Oort, which tend to prioritize high-loss clients—enhancing robustness in the presence of noisy data.

\begin{figure}[h]
    \centering
    \includegraphics[width=0.42\textwidth]{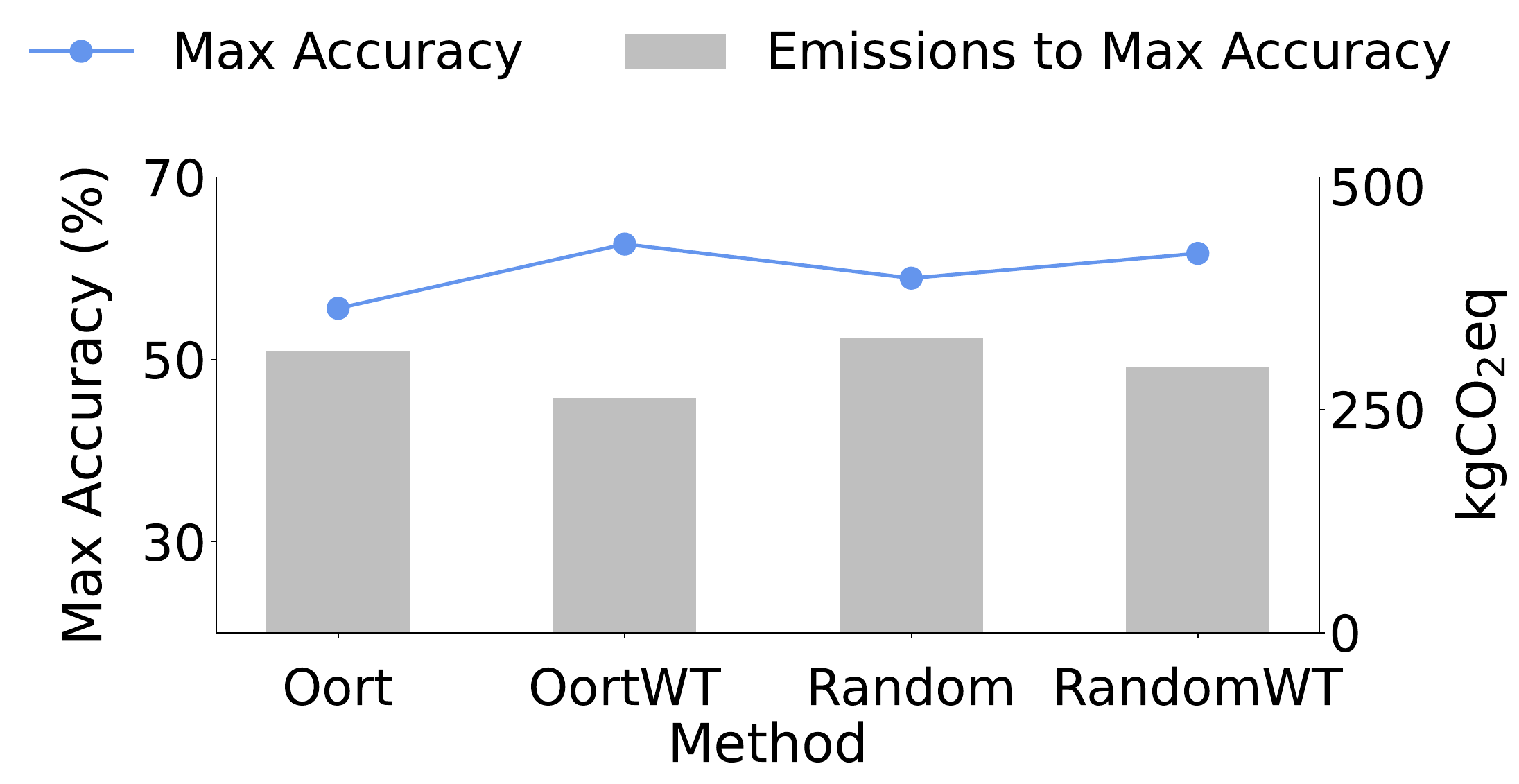}
  \caption{Accuracy–emissions comparison across methods. Bars show the carbon emissions spent during training up to the point of maximum accuracy, while the line shows the corresponding maximum accuracy.}
  \label{fig:emtoacc}
\end{figure}

While the average carbon intensity of corrupted clients is lower than that of others (see client IDs 0–5 in Fig.\ref{fig:carbonintensity}), their selection leads to increased carbon emissions while achieving lower accuracy. Although the probing round introduces a one-time computational cost, it allows reaching maximum accuracy earlier, allowing for a reduction in training rounds, as demonstrated in Figure \ref{fig:emtoacc}, which shows emissions measured at the point of maximum accuracy.

\subsection{Evaluating Budget-Constrained Utility-Aware Client Selection}

\begin{figure}[h]
  \centering
    \centering
    \includegraphics[width=0.98\linewidth]{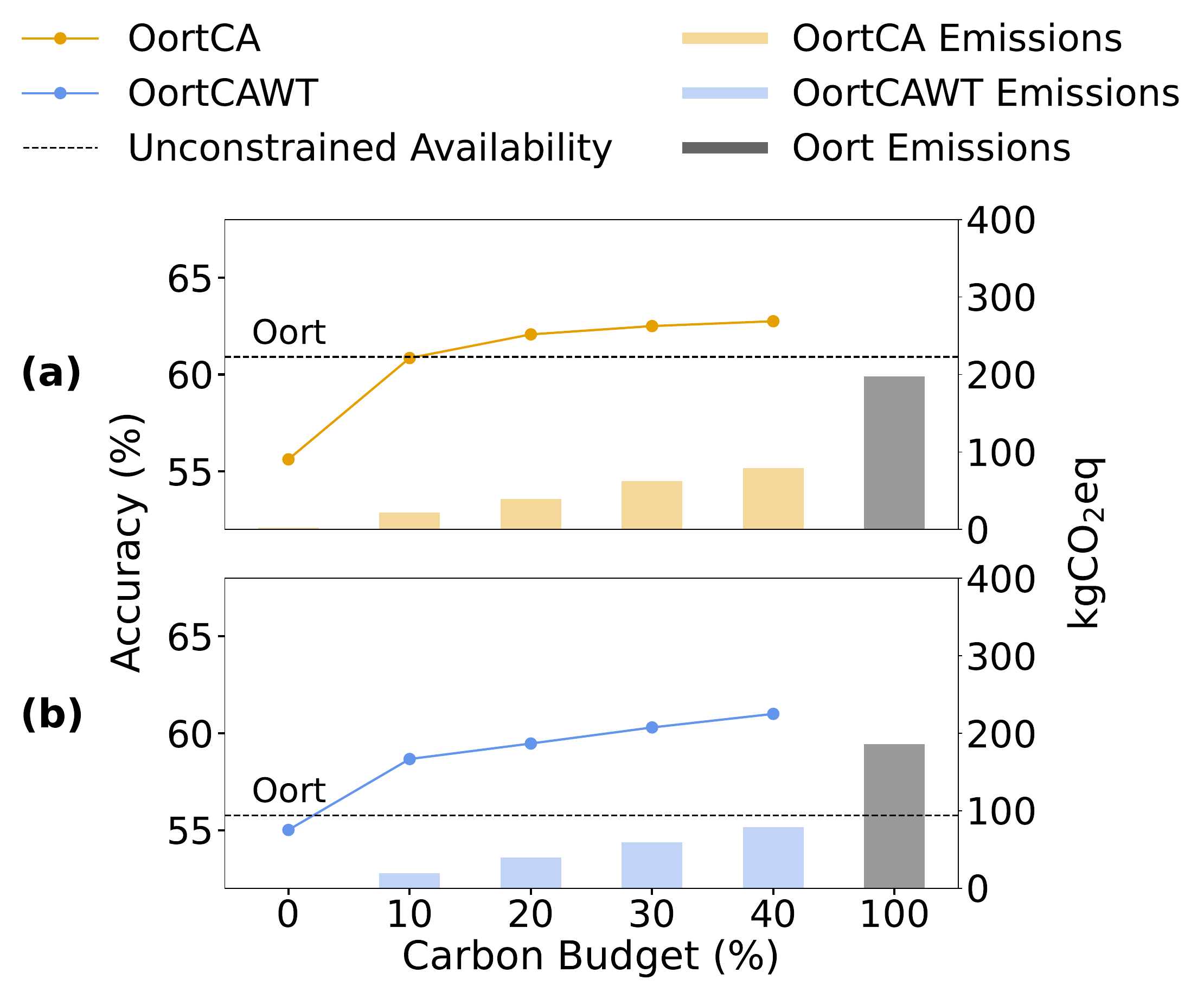}
    \caption{Impact of carbon budgeting on CO$_2$-emissions and model performance, compared to standard Oort client selection with full client availability. Results are shown for two scenarios: (a) clients with clean data, and (b) clients with corrupted data.}
    \label{fig:accuracy_vs_budget}    
\end{figure}

\begin{figure*}[h!]
    \centering
    \includegraphics[width=0.9\textwidth]{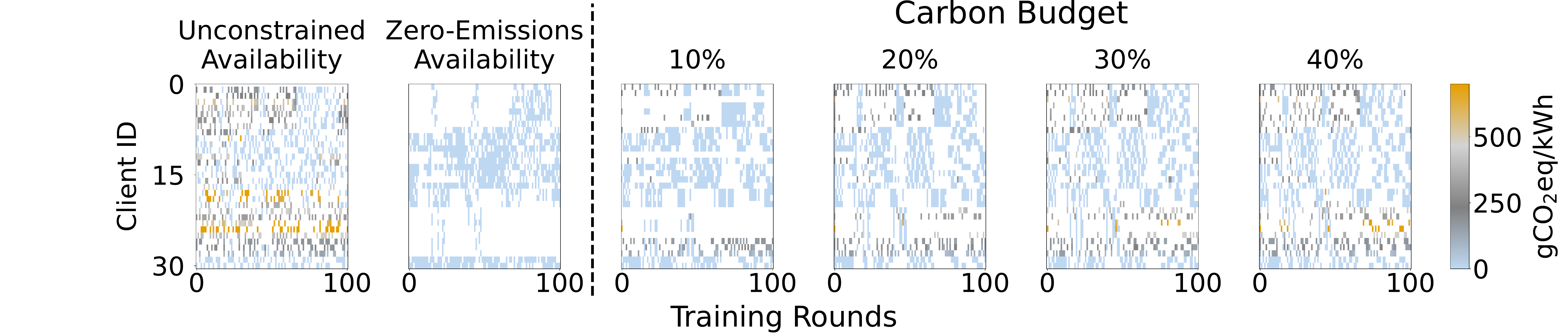}
    \caption{Client selection during training: Comparison between two approaches—one with unconstrained emissions allowing full client participation in each round, and another enforcing a hard constraint where only zero-emission clients are selected. A third, more flexible strategy, balancing carbon budgeting with broader client availability during training.}
    \label{fig:budget-selection}
\end{figure*}

We evaluate the utility-aware carbon budget allocation strategy introduced in Section~\ref{sec:oortca}, here referred to as OortCA. We compare it to the original Oort strategy, which serves as the baseline with unconstrained emissions usage—meaning \textit{all clients are available} for selection regardless of their carbon intensity. The total emissions from Oort serve as a reference point for setting carbon budgets and define the emission baseline.

Carbon budget is set for the entire training and divided evenly across rounds, with unused carbon carried over. If the budget is spent before reaching the desired number of clients in a round, as the remaining are selected curtailment clients with the highest utility scores. 

We test OortCA across different levels of carbon budget relative to this emission baseline. Evaluation proceeds as follows: we begin with a zero-carbon budget, which restricts selection to clients using only curtailed (i.e., zero-emission) energy, significantly limiting the available client pool. We then increase the carbon budget in 10\% increments, gradually expanding the set of eligible clients for selection and allowing higher-emission clients to be included, as shown in Figure~\ref{fig:budget-selection}.

As shown in Figure~\ref{fig:accuracy_vs_budget}, by effectively balancing client utility despite reduced availability due to budget constraints, OortCA achieves final accuracy comparable to the unconstrained Oort baseline — while using only 40\% of its emissions.

We further repeat the experiment under the noisy data setup, combining the utility-aware budget allocation (Section~\ref{sec:oortca}) with the gradient norm thresholding approach (Section~\ref{sec:oortwt}). The resulting method, referred to as OortCAWT, enables more robust client selection in the presence of corrupted data while maximizing the use of limited carbon budgets and spending the budget on clean, high-utility clients.

\begin{figure*}[h!]
  \centering
    \centering
    \includegraphics[width=0.78\linewidth]{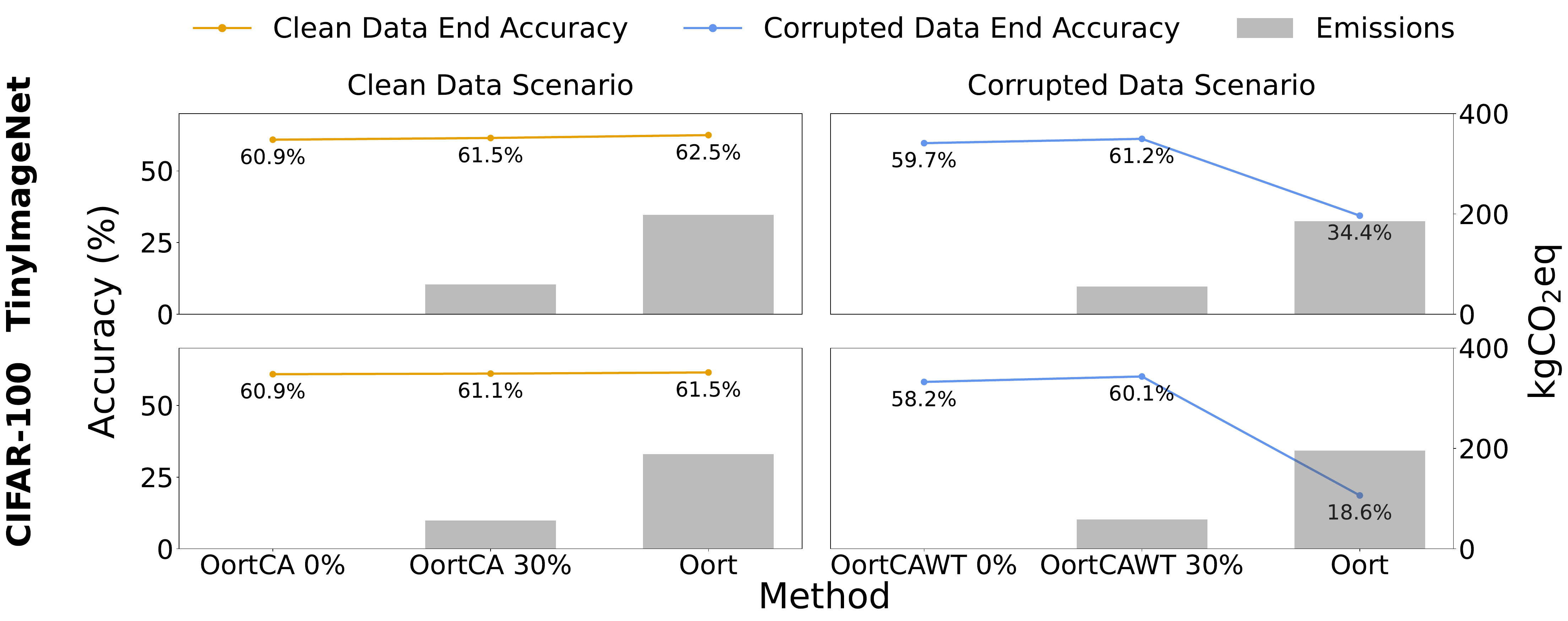}
    \caption{Impact of carbon budgeting on CO$_2$-emissions and model performance using the DenseNet-121 model evaluated on the CIFAR-100 dataset and the EfficientNet-B1 model evaluated on the Tiny ImageNet dataset. Results are shown for two scenarios: clients with clean data and corrupted data, where noise was added to 6 out of 30 clients.}
    \label{fig:cifar100}
\end{figure*}

Figure \ref{fig:cifar100} presents extended results using the DenseNet-121 \cite{huang2017densely} model evaluated on the CIFAR-100 dataset and experiments with the EfficientNet-B1 \cite{tan2019efficientnet} evaluated on the Tiny ImageNet. The Tiny ImageNet contains 100,000 64×64 color images of 200 classes. CIFAR-100 \cite{krizhevsky2009learning} contains 60,000 32x32 color images across 100 classes.

The results suggest that in a clean data scenario, the client selection strategy manages to archieve similiar model performance even with a reduced/constrained client availability. While in under noisy data conditions model performance varies drastically. In this case, filtering out noisy data via gradient norm improves robustness. Moreover, when combined with carbon-aware training (e.g., budgeted carbon emissions), this approach enables accuracy improvements while accounting for reduced carbon emissions. This opens the possibility of strategically allocating carbon budgets to achieve accuracy gains in noisy data scenarios.

\section{Future Work}
\label{sec:discussion}

In this section, we outline potential future directions informed by the experiments and insights gained from this work. 

\textbf{Data Valuation in Carbon-Aware Federated Learning}  
FL enables privacy-preserving distributed training, offering a platform to explore data valuation methods for sustainable AI. Techniques such as Federated Shapley Values and Gradient Alignment should be evaluated in carbon-aware settings to balance fairness, bias mitigation, and carbon efficiency \cite{NEURIPS2021_8682cc30, wang2020principled}. Asynchronous Federated Learning (AFL) is a promising alternative, removing reliance on potentially inaccurate and costly carbon intensity forecasts. Coupling AFL with data valuation and carbon budgeting may yield practical, carbon-efficient training strategies. Another direction involves leveraging critical learning periods—stages where models are most sensitive to high-quality data. Aligning these phases with moderately carbon-intensive, high-value data could improve both convergence and sustainability \cite{achille2017critical}. To reduce computation during probing rounds, data coresets can be estimated via a single round of local inference \cite{mirzasoleiman2020coresets}. This lowers the cost of gradient norm calculations and can aid in efficiently onboarding new clients.

\section{Conclusion}
\label{sec:conclusion}

In this paper, we investigated the impact of noisy data on client selection strategies and their implications for carbon-aware computing. We also explored the influence of carbon budgeting for larger client availability and its effect on model performance. Our experiments demonstrate that gradient norm thresholding is effective for filtering noisy clients, and that strategic use of carbon budgets can mitigate the effects of sparse client availability during periods of renewable energy volatility. We hope this work not only advances understanding in this area but also encourages further research into practical and efficient approaches to carbon-aware machine learning.


\bibliographystyle{ACM-Reference-Format}
\bibliography{sample-base}

\end{document}